\documentclass[runningheads]{llncs}

\usepackage{eccv}

\usepackage{eccvabbrv}

\usepackage{graphicx}
\usepackage{booktabs}

\usepackage[accsupp]{axessibility}  

\usepackage{algorithm}
\usepackage{algpseudocode}
\usepackage{graphicx}
\usepackage{subcaption}
\usepackage{multirow}
\usepackage{xcolor}
\usepackage{colortbl}
\usepackage{array}

\usepackage[pagebackref,breaklinks,colorlinks,citecolor=eccvblue]{hyperref}

\usepackage{orcidlink}

\def\ie{\textit{i.e.}}
\def\eg{\textit{e.g.}}

\definecolor{color3}{rgb}{0.95,0.95,0.95}
\begin{document}

\title{Towards Scalable, Flexible, and Adaptive Multi-Modal Face Synthesis} 

\titlerunning{Abbreviated paper title}

\author{Jingjing Ren\inst{1} \and
Cheng Xu\inst{2}\and
Haoyu Chen\inst{1}\and
Xinran Qin \inst{3}\and
Lei Zhu \inst{3}
}

\authorrunning{F.~Author et al.}

\institute{
The Hong Kong University of Science and Technology (Guangzhou) \and
Centre of Smart Health, The Hong Kong Polytechnic University
\and
School of Cyber Science and Technology, Shenzhen Campus of Sun Yat-sen University
\\
\url{https://jingjingrenabc.github.io/multimodal-face-synthesis/} 
}

\maketitle

\begin{center}
    \centering
    \includegraphics[width=1\textwidth]{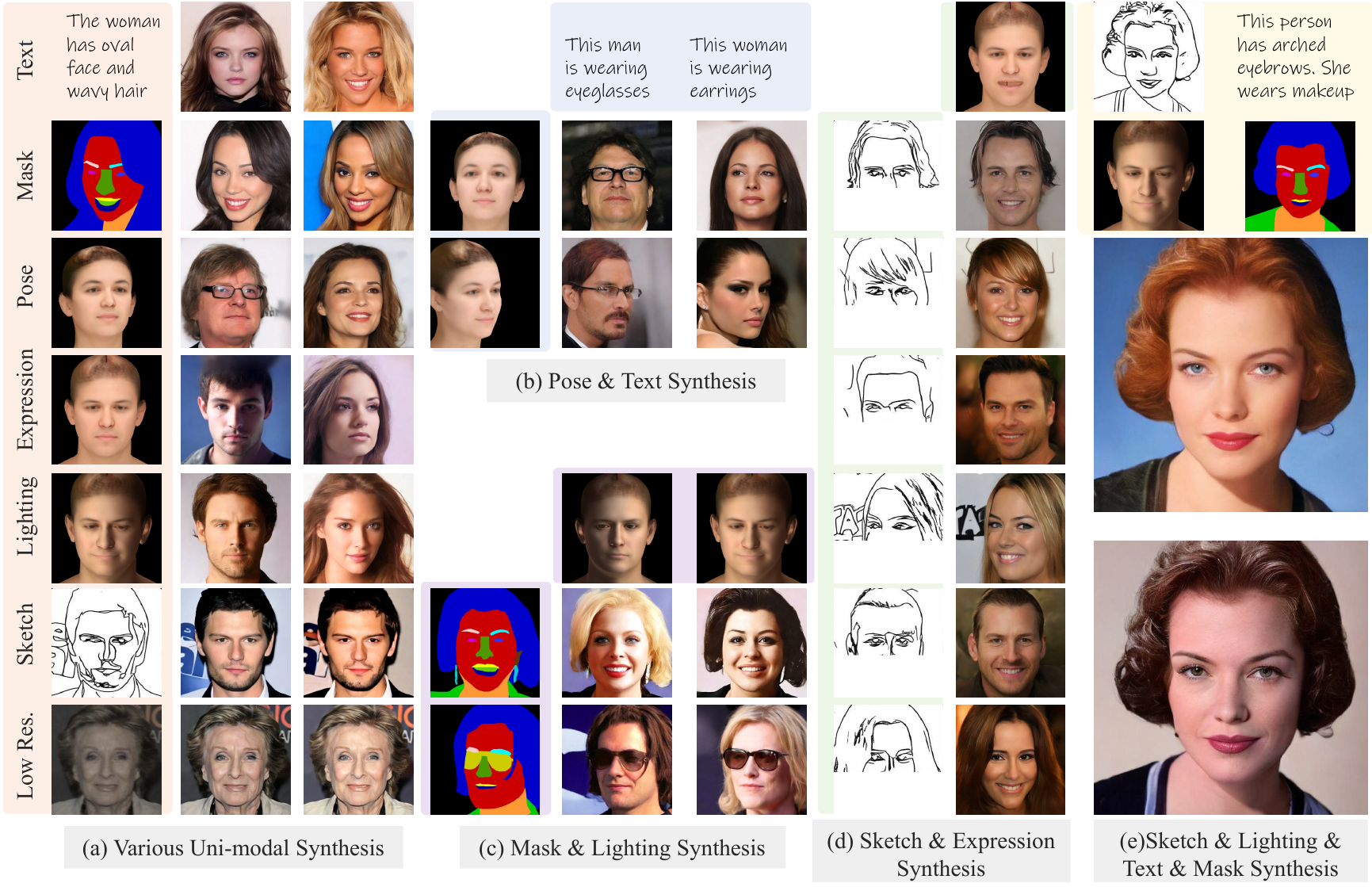}
    \vspace{-8mm}
    \captionof{figure}{Our method's versatile synthesis capabilities, demonstrating high-fidelity facial image generation from a flexible combination of modalities. Remarkably, these diverse face synthesis tasks are achieved within a single sampling process of a unified diffusion U-Net, demonstrating the method's efficiency and the seamless integration of multi-modal information.
    } 
    \label{fig:teaser}
   \end{center}%

\begin{abstract}
  Recent progress in multi-modal conditioned face synthesis has enabled the creation of visually striking and accurately aligned facial images. Yet, current methods still face issues with poor scalability, limited flexibility, and a one-size-fits-all approach to control strength for various conditions. To address these challenges, we introduce a novel uni-modal training approach with modal surrogates, coupled with an entropy-aware modal-adaptive modulation, to support flexible, scalable, and adaptive multi-modal conditioned face synthesis. Our  modal surrogate decorate condition with modal-specific characteristic and serve as linker for inter-modal collaboration, resulting in a highly scalable and flexible multi-modal face synthesis framework. The entropy-aware modal-adaptive modulation finely adjust diffusion noise according to modal-specific characteristics and given conditions. It enables well-informed step along de-noising trajectory and ultimately leads to synthesis results of high fidelity. Building upon our scalable, flexible and adaptive multi-modal synthesis framework, we efficiently incorporate more modalities and support a wide range of face synthesis applications. Our extensive experiments demonstrate our method's superiority for multi-modal face synthesis.
  \keywords{Multi-modal face synthesis \and Diffusion model }
\end{abstract}
\section{Introduction}
Diffusion models have made remarkable achievements on a wide range of synthesis tasks in various domains e.g. image ~\cite{rombach2022high, kim2022diffusionclip}, audio  \cite{pascual2023full, kong2020diffwave} ,video \cite{blattmann2023align, esser2023structure} and motion \cite{zhang2022motiondiffuse, yuan2023physdiff}.
Within the sphere of image synthesis, the focus has increasingly shifted towards more controllable synthesis under multi-modal conditions \cite{zhang2023adding, mou2023t2i, nair2023unite, liu2022compositional, ham2023modulating, zhao2024uni, qin2023unicontrol}. These models can generate images that are not only visually compelling but also align well with given multi-modal conditions.
The emerging developments in this field are setting the stage for more dynamic and user-centric image synthesis, broadening its practicality and impact in real-world applications.

\begin{figure}[t] 
    \centering 
    \includegraphics[width=\linewidth,height=0.2\linewidth]{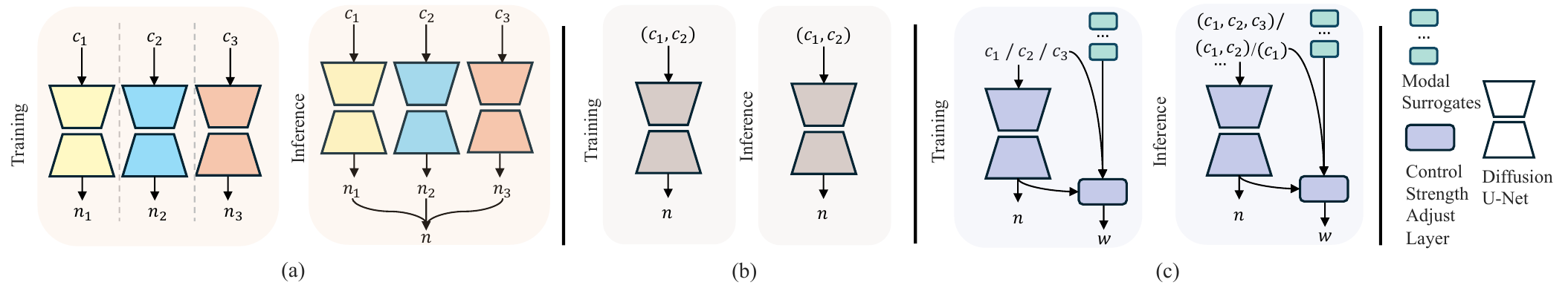} 
    \caption{Core idea comparison between existing multi-modal synthesis approaches and our method. (a) Fusing noises from multiple uni-modal diffusion models. (b) Incorporation of additional control mechanisms in basic synthesis models for multi-modal synthesis conditioned synthesis. (c) Our method achieve multi-modal conditioned face synthesis within a single synthesis network, under flexible combination of conditions and dynamically adjust noise of diffusion step.} 
    \label{fig:overview_compare} 
\end{figure}

One approach to achieve multi-modal conditioned synthesis, as explored in recent studies \cite{nair2023unite, liu2022compositional, huang2023collaborative}, involves first developing uni-modal condition synthesis diffusion models and then fusing the noises generated by each modal branch, as shown in Fig.~\ref{fig:overview_compare} (a). This technique allows for flexible multi-modal conditions synthesis, enabling image generation under any combination of modal inputs. However, a notable limitation is its limited scalability. Each modality requires a distinct synthesis network, and the inference complexity increases linearly with the involvement of new modality. 
%
Another alternative approach \cite{qin2023unicontrol, mou2023t2i, zhang2023adding, zhao2024uni} to multi-modal conditioned synthesis involves incorporating additional control mechanisms into basic synthesis models, as shown in Fig.~\ref{fig:overview_compare} (b). 
These innovations enable image synthesis under the combined modal conditions of layouts and text. However, these methods fall short of flexibility. They rely on specific tuning for each unique combination of modalities and require multi-modal annotated data.

\begin{figure}[t] 
    \centering 
    \includegraphics[width=\linewidth]{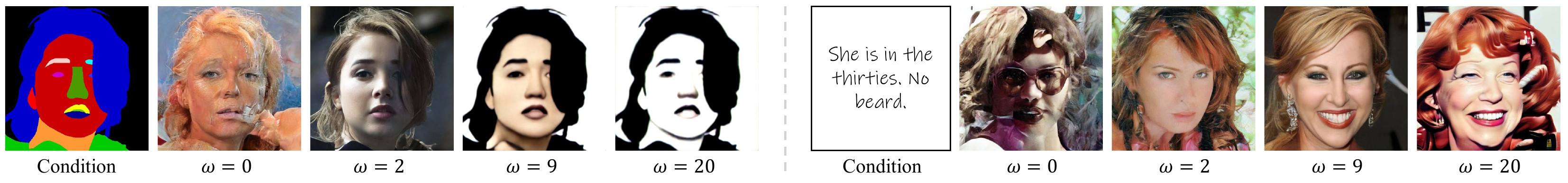} 
    \caption{Uni-modal synthesis results given different control
strength $w$. To generate facial images of high fidelity and quality, text and mask require different control strength due to their entropy difference.} 
    \label{fig:introw} 
\end{figure}

Furthermore, conditions from different modality naturally exhibits diverse condition entropy, a measure of unpredictability in data given some condition.
A condition of higher entropy requires higher control strength to exert adequate influence on face synthesis, while some condition of lower entropy suffer from over-fitting with the same control strength.
As shown in Fig.~\ref{fig:introw} where higher $w$ indicate higher control strength,  a high-entropy modality (text), benefits from stronger control, while masks with lower entropy risk over-fitting. 
Existing multi-modal conditioned synthesis~\cite{nair2023unite, liu2022compositional, huang2023collaborative, zhao2024uni, qin2023unicontrol} often neglect such inherent differences in conditional entropy for various modalities.
They consequently assign equal control across modalities, and produce results of poor alignment with given multi-modal condition.
%
%

To address the limitations of existing methods in terms of scalability, flexibility and adaptivity, we introduce a uni-modal training approach with modal surrogate for enhanced flexibility and scalability, coupled with an entropy-aware modal-adaptive modulation mechanism for responsive adaptation to varying modal conditions. 
As shown in Fig.~\ref{fig:overview_compare}, we set a distinct modal surrogate for each modality, designed to function as both a condition decorator for its respective modality and an inter-modal linker to facilitate collaboration between different modalities. 
During training, we only use uni-modal annotated data. 
The surrogate of the active modality is merged with the given condition, capturing modality-specific context as a complement to the given conditions.
On the other hand, surrogates of other modalities is also involved with the given condition, to learn collaboration with the activate modality.
With the surrogate decorating function and inter-modal learning, the resulting network can discriminate and process condition among different modalities.
Therefore our synthesis network is highly flexible and scalable, capable of generating facial images under a variety of modal combinations within a single sampling process of a single diffusion model.
To take into full consideration the disparate conditional entropy inherent to different modalities, we further develop an entropy-aware modal-adaptive modulation mechanism. 
By thoughtfully adjusting the noise levels, this mechanism allows the network to adapt its de-noising strategy to the distinct characteristics of each modality. Consequently, this adaptive approach ensures that each modality sufficiently guides the image generation process, leading to faithful synthesis to all given conditions.

Using our novel approach of uni-modal training with modal surrogates, equipped with by an entropy-aware, modal-adaptive modulation mechanism, our framework excels in generating high-fidelity facial images across flexible combination of conditions. 
We extend beyond the standard modalities of masks, text and sketch ever considered in existing methods \cite{huang2023collaborative, nair2023unite, liu2022compositional, xia2021tedigan, zhang2023adding, mou2023t2i} for face synthesis, incorporating lighting, pose, expression and low-resolution images to enrich the multi-modal synthesis applications. This allows for a wide range of applications, \eg, various uni-modal face synthesis in Fig. \ref{fig:teaser} (a), pose and text conditioned face synthesis in Fig. \ref{fig:teaser} (b), lighting and sketch conditioned face synthesis in Fig. \ref{fig:teaser} (d), 
and so on.

Our contributions are summarized as following:
\begin{itemize}
    \item We devise an uni-modal training framework that assigns a unique modal surrogate to each modality, greatly enhancing the face synthesis process in terms of flexibility and scalability. These modal surrogates serve dual purposes: acting as condition decorators for their respective modalities and as inter-modal linkers, enabling our network to efficiently process a wide array of modal inputs within a unified diffusion model.
    \item  We propose an entropy-aware modal-adaptive modulation that dynamically adjusts the noise levels based on the conditional entropy of each modality. This allows our system to finely adjust the de-noising process, ensuring effective utilization of specific information from each modality and resulting in face synthesis of high fidelity and quality.
    \item Our framework incorporates conditions of more modalities to significantly broaden the scope of our multi-modal face synthesis capabilities, supporting a diverse range of face synthesis under uni-modal and complex multi-modal condition. This demonstrates the versatility and creative potential of our approach.
\end{itemize}
\section{Related Work}
\subsection{Latent Diffusion Model}
Recent advancements in image synthesis have been significantly driven by the development of diffusion models, which have shown remarkable success across various domains of generative tasks \cite{rombach2022high, kim2022diffusionclip, dhariwal2021diffusion, gu2022vector, ding2023diffusionrig, yuan2023physdiff, zhang2022motiondiffuse}. 
Among these, latent diffusion models \cite{rombach2022high} stand out as a powerful subclass that operates on compressed representations of data. 
These models map high-dimensional data into a latent space where the diffusion process is applied, leading to efficient synthesis with reduced computational demands. %
The compressed latent representations retain essential information while filtering out noise, enabling the models to focus on generating coherent structures in the image synthesis process. 
This approach has opened up new avenues for creating high-quality images that are both diverse and reflective of complex conditional inputs, setting new benchmarks in the field of generative modeling.
We build our multi-modal conditioned face synthesis network upon latent diffusion framework.
 \vspace{-2mm}
\subsection{Conditioned Face Synthesis}
 \vspace{-2mm}
In the evolving landscape of conditioned face synthesis, researchers have explored a range of controls to guide the synthesis process. This includes face generation under condition of sketches \cite{xia2021cali, ghosh2019interactive}, semantic masks \cite{park2019semantic}, 3D face models \cite{chan2022efficient, chan2021pi, tan2022volux, ding2023diffusionrig}, low-resolution images \cite{wang2018esrgan}, and textual descriptions \cite{reed2016generative, tao2020df, zhang2017stackgan, liu2022controllable}.
Recent methodologies \cite{xia2021tedigan, nair2023unite, liu2022compositional,huang2023collaborative, mou2023t2i, zhang2023adding, tang2023any, huang2022multimodal} have introduced multi-modal conditions into the face synthesis framework to achieve rich and flexible control than uni-modal condition. 

One line of methods \cite{huang2023collaborative, nair2023unite, liu2022compositional} sample multi-condition distributions (Eq.~\ref{eq:cond_prob}) using stochastic gradient Langvein sampling, converting uni-modal classifier-free guidance (CFG)\cite{ho2022classifier} for diffusion model  into a multi-conditioned version:
{
\begin{equation}
\label{eq:cfg}
 \epsilon(x_t,t,C)=  \sum_{m=1}^{M}(w_m+1)\epsilon_{\theta_m}(x_t, t, c_m) -  \sum_{m=1}^{M} w_m \epsilon_{\theta_m}(x_t, t).
\end{equation}}where $C=\{c_1,...,c_M\}$ respresents multi-modal condition and $w_m$ denotes the controlling strength of the $m$-th modality.
As defined in Eq.~\ref{eq:cfg}, each modality $m$ employs an individual synthesis network, ignoring the shared aspects of image generation.
This technique allows for flexible synthesis across modalities. However, it tends to overlook the shared synthesis properties across different modalities, and suffer from an increased sampling complexity proportionate to the number of modalities.

A concurrent technological trajectory~\cite{zhang2023adding, mou2023t2i, zhao2024uni, qin2023unicontrol} achieve multi-modal conditioned synthesis by adding layout control to a pre-trained text-guide image synthesis model, which can be presented as:
{
\begin{equation}
\begin{split}
\label{eq:cfg_control}
 \epsilon(x_t,t,C_{l}, c_{text})=  (w + 1)\epsilon_{\theta}(x_t, t, C_l, c_{text}) -
 w\epsilon_{\theta}(x_t, t, C_l),
\end{split}
\end{equation}}where $C_l$ is the layout condition of a single or multiple modalities. $w$ represents the controlling strength of the textual modality.
As shown in Fig.~\ref{fig:overview_compare} (b), these methods use a parameter-sharing network for multiple modalities guided synthesis. 
However, they directly integrate controlling features of different modalities for multiple layout conditioned synthesis, tending to produce inferior results due to the lack of inter-modal collaboration during the model learning.
While this approach processes multiple modal inputs within a single network, it lacks flexibility and often requires additional training for each combinations of modalities.
%
%
Our method integrate the distinct characteristics of each modality and interaction among them within a unified, adaptive synthesis framework to support multi-modal conditioned face synthesis.

\section{Method}
\begin{figure}[!tbp] 
    \centering 
    \includegraphics[width=0.75\linewidth]{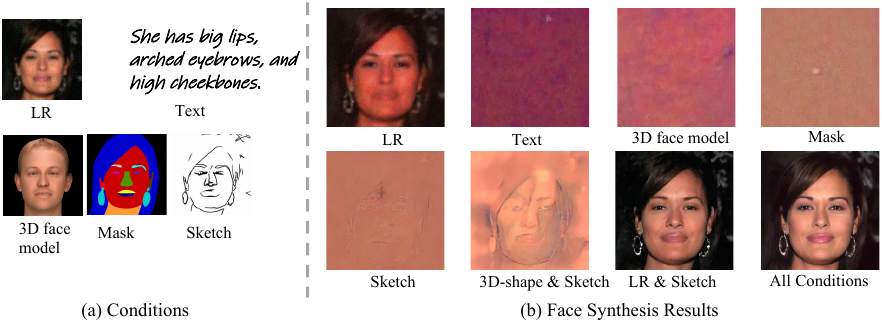} 
    \caption{Results of multi-modal training. The left are input multi-modal conditions. The synthesis results are presented in the right part. The resulting network can only synthesize pleasing results given all conditions and much of the guidance comes from the low-resolution image. The synthesis network tend to rely on modality of low condition entropy (LR) for synthesis and thus neglect modality with higher condition entropy.} 
    \label{fig:multires} 
\end{figure}

\begin{figure*}[t]
    \centering
    \begin{minipage}[c]{0.44\textwidth}
        \includegraphics[width=\textwidth]{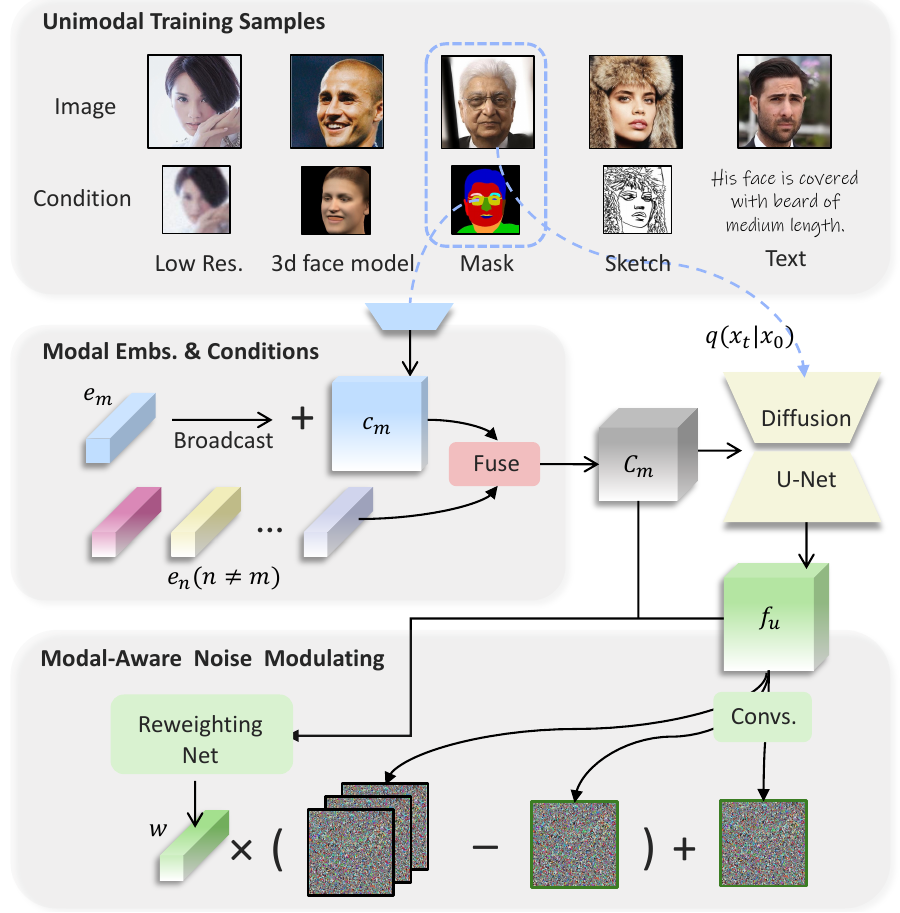} 
    \end{minipage}
    \hfill
    \begin{minipage}[c]{0.52\textwidth}
        \vspace{-3mm}
        \begin{algorithm}[H]
            \scriptsize
            \caption{Uni-modal Training with Modal Surrogates}
            \label{algo:unimodal}
            \begin{algorithmic}[1]
                \Repeat
\State \( (x_0, c_m),  m \sim \mathcal{U}(\{1,\dots, M\})\) 
\State    \Comment{\textcolor{gray}{Sample from uni-modal datasets }}
\State \( t \sim \mathcal{U}(\{0,\dots, T\}), \epsilon \sim \mathcal{N}(0, \mathcal{I}) \)
\State \( x_t = \sqrt{\Bar{\alpha_t}}x_0 + \sqrt{1 - \Bar{\alpha_t}}\epsilon  \)
\State \(C_m = f(c_m, e_1,\dots, e_M)\)
\State \Comment{\textcolor{gray}{Fuse condition and all the modal surrogates.}}
\State \( L = ||  \epsilon_{\theta_u, \theta_w}( x_t, C_m, t) - \epsilon ||^2 \) 
\State \(\theta_* \leftarrow \theta_* - \eta \frac{\partial L}{\partial \epsilon_{\theta_*}}\frac{\partial \epsilon_{\theta_*}}{\partial \theta}, * \in \{u, w\}\)
\State \(e^n \leftarrow e^n - \eta \frac{\partial L}{\partial \epsilon_\theta}\frac{\partial \epsilon_\theta}{\partial C^m}\frac{\partial C^m}{\partial e^n} \)
\State \Comment{\textcolor{gray}{Updating surrogate of ALL modality}}
\Until{converged}
            \end{algorithmic}
        \end{algorithm}
    \end{minipage}
        \caption{Uni-modal training with modal surrogates and entropy-aware modal-adaptive modulation mechanism. During training, we randomly sample uni-modal data, of which the condition is fused with its modal surrogate to learn modal-specific intrinsic and other modal surrogate to learn inter-modal collaboration. The fused features are sent to the diffusion U-Net to guide the de-noising process of the corrupted input image. The output noise is further modulated according to the condition features and UNet feature to adaptively adjust noise level given the conditions. The training process is provided in Algorithm~\ref{algo:unimodal}.}
        \label{fig:unimodal}
\end{figure*}


The goal to synthesize facial image $x$ given conditions from multiple modalities, \eg, text and mask,
can be formally formulated as learning:
{
\begin{equation}
\label{eq:cond_prob}
    P(x|c_1, \dots, c_M),
\end{equation}}where $c_m$ denotes the $m$-th condition modalities and $M$ is the number of condition modalities.
%
%
we devise an efficient, flexible, and adaptive multi-modal conditioned face synthesis, as shown in Fig.~\ref{fig:overview_compare}(c).
In the following sections, we first elaborate in detail our uni-modal training via modal surrogates for efficient and flexible synthesis (Sec.~\ref{sec:eff_method}). Afterwards, we present our adaptive weighting mechanism for adaptive synthesis (Sec.~\ref{sec:reweight}).

\subsection{Uni-modal Training with Modal Surrogates}
 \label{sec:eff_method}
To ensure efficient synthesis, we adopt a shared synthesis network for conditions from all modalities as previous works~\cite{mou2023t2i, zhang2023adding, qin2023unicontrol, zhao2024uni}.
This network is required to discriminate between those different modalities, to support various conditioned face synthesis within a single network. 
To this end, we assign a modal surrogate to each modality, for decorating the condition with its modal-specific information. 
The learning objective can be represented as 
\begin{equation}
\label{eq:uniloss1}
L := \mathbb{E}_{t, x^i_0, \epsilon\sim\mathcal{N}(0, 1)}[|| \epsilon_\theta(x^i_t, f(c^i_m, e_M), t), \epsilon||_2],
\end{equation}
where $x^i_t$ is noised version of the $i$-th input image $x^i_0$. $f(\cdot)$ denotes the fusion operation of condition $c^i_m$ and the modal surrogate $e_m$ of its modality. 
The modal surrogate is a learnable token. We instantiate $f(\cdot)$ by broadcasting the modal surrogate and adding it to the condition $c_m$.
%
The modal surrogate is to decorate condition with modal intrinsic and helps network discriminate among different modal.
Thus far, our model is capable of synthesizing faces given various single modal condition, \eg, mask, sketch or text.
%

%

To achieve multi-modal face synthesis, a straightforward way is to train a unified model with paired multi-modal annotated data (\ie, $(x^i, c_1^i,...c_M^i)$) to generate faces that align with the multi-modal conditions simultaneously.
However, it would be extremely complex and tedious to consider all those combinations during training as the total number of combinations of $M$ modalities could be $C_M^1 + C_M^2 + ... + C_M^M$. 
Moreover, in multi-modal training, the network may prioritize the most informative condition to loosen the learning difficulty. To demonstrate this, we illustrate a representative example where network relies on the low resolution image condition most to render the face, while neglecting other modalities in Fig.~\ref{fig:multires}. 
To solve the above issue, we design a inter-modal learning  mechanism that enables the network to not only learn form uni-modal data $x^i, x^i_m$ as in Eq.~\ref{eq:uniloss1}, but also take fully advantage of rich modality cues from modal surrogates of other modalities. 
Therefore, the training objective in Eq.~\ref{eq:uniloss1} can be then reformulated as: 
\begin{equation}
\label{eq:uniloss2}
L := \mathbb{E}_{t, x_0, \epsilon\sim\mathcal{N}(0, 1)}[|| \epsilon_\theta(x^i_t, f(c^i_m, e_1, ..., e_M), t), \epsilon||_2], 
\end{equation}
where the modal surrogates of other modalities are also involved when data of condition $c_m$ is fed for training.
Note that the modal surrogates of modal $c_m$ in Eq.~\ref{eq:uniloss2} is similarly defined as in Eq.~\ref{eq:uniloss1}, but it additionally learns inter-modal interaction with other modals.
For modal surrogates of other modals $n$ ($n \neq m$), their modal surrogates also get updated.
In this way the surrogates of some modal also learn from data of other modalities, and thus effectively capture inter-modal interactions with modal $m$ during the whole training process. 
Such inter-modal modelling helps aid network in grasping inter-modal collaboration via uni-modal training only.
%
%
Consequently our method can perform flexible synthesis given arbitrary combination of condition modalities within a single sampling process of a single synthesis network.

\subsection{Entropy-Aware Modal-Adaptive Modulation}
\label{sec:reweight}
Multi-modal conditions naturally exhibits diverse entropy.
Therefore it is essential to finely adjust the de-noise strategy according to multi-modal condition and modal-specific characteristic, ensuring generating vivid and faithful facial images.
%
%
For uni-modal synthesis, one can flexibly specify the controlling strength to achieve high-quality synthesis results.
However, in the multi-modal synthesis scenario, it is not desirable to employ a fixed controlling strength across all modalities. The reason behind this is that for some low controlling strength, the conditions of high condition entropy may have insufficient influences on image synthesis process. 
Simply increasing controlling strength may lead to overfitting of the generated images to some other condition of low condition entropy. 
%
We devise an entropy-aware modal-adaptive modulation to adaptively adjust the predicted noise according to given conditions and current de-noising state.
As revealed in Fig.~\ref{fig:unimodal}, we use a weighting module that processes the fused features $C_m$ of conditions and modal surrogates, and feature $f_u$ of the final decoding layer in U-Nnet. This network performs average pooling on these features, and then passes them through several linear layers to produce a weighting vector $w$.
Note that the modal surrogates encapsulate modality-specific priors, and therefore they offer critical guidance on noise adjustment when combined with the given conditions. The inclusion of U-Net features also injects real-time insights $f_u$ into the de-noising trajectory.
Apart from predicting a single base noise map based on the U-Net feature $f_u$, we follow the philosophy of multi-head \cite{vaswani2017attention} and predict multiple noise maps, each potentially capturing different aspects of the features.
Finally, we can obtain the output noise that can be presented as:
{
\begin{equation}
\begin{split}
\label{eq:final_noise}
\epsilon_\theta=\frac{1}{K} \sum_k ( w_k(n_k-n_b) + n_b),
\end{split}
\end{equation}
}where $k$ denotes the number of additionally predicted noise maps.
We start with a base noise pattern $n_b$, and introduce adjustment from other noise maps $n_k$. The influence of$n_k$ is scaled by $w_k$ according to modal-specific information, given conditions, and current state. 
This mechanism enables well-informed and strategic decisions in each step of de-noising, leading to the final high-quality synthesis aligned well with multi-modal conditions.

\section{Experiments}
In this section we first introduce our experimental settings in Sec. \ref{sec:expsetup}. Then we show various face synthesis applications based on our framework in Sec. \ref{sec:application}. Next we conduct comparative study over existing competitive methods in Sec. \ref{sec:compare_study}. Finally we demonstrate effectiveness of our method via ablation analysis in Sec. \ref{sec:abla_anal}.
\subsection{Experimental Setup}
\label{sec:expsetup}
\textbf{Datasets.} 
We conduct experiments on the Celeb-HQ dataset \cite{karras2017progressive} with 30k high-quality face images.
We utilized mask from \cite{lee2020maskgan} and textual descriptions from \cite{jiang2021talk}. We compared our method, focusing on parsing mask and text modalities, for fair comparison with existing multi-modal methods following TediGAN \cite{xia2021tedigan} and collaborative Diffusion\cite{huang2023collaborative}. 
We train our method on the first 27K images and 
Our approach further efficiently includes additional modalities such as sketches \cite{xia2021tedigan}, 3D face models \cite{feng2021learning}, and low-resolution images \cite{wang2021towards}, enriching our range of synthesis tasks.

\noindent
\textbf{Implementation details.} 
We implement our method based on Latent Diffusion \cite{rombach2022high}, where the images $(3\times H\times W) $are first mapped to latent features $(3 \times \frac{H}{8} \times \frac{W}{8})$.
The modal encoder of the low-resolution image is the same as the first stage encoder of LDM \cite{rombach2022high}.
The encoder of text and low-resolution image is fixed during training, while the encoders of other modalities are updated in training.
The modal surrogate for text is initialized as the textual encoder feature of input \textit{human face} and others are randomly initialized.
We train our method for 40k iterations on four RTX3090 GPUs with batch size of 32.
\textit{We shall release our code upon publication of this manuscript.}

\noindent
\textbf{Evaluation metrics.} 
We assess our method's performance using Frechet Inception Distance (FID) for image quality and diversity, text matching accuracy through CLIP's vision-text space cosine similarity (measuring alignment between generated images and text descriptions), and mask accuracy, evaluated by comparing parsing results of generated images against ground-truth semantic maps.

\begin{figure}[t] 
    \centering 
    \includegraphics[width=1\linewidth]{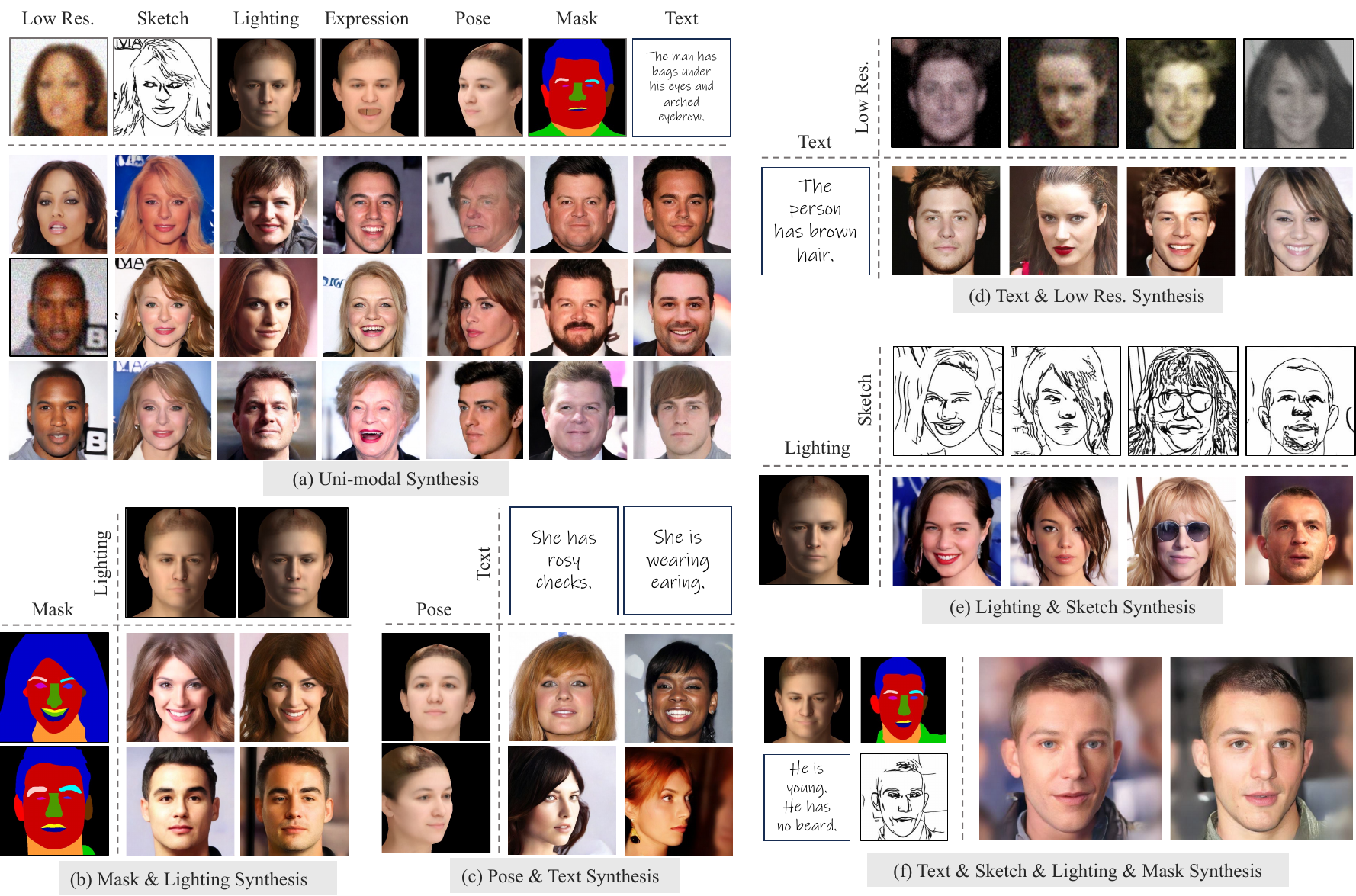}     \caption{ Demonstrating synthesis capability under the combination of any modalities, including uni-modal conditioned synthesis, pose \& text conditioned synthesis, and text \& sketch \& lighting \& mask conditioned synthesis, etc.} 
    \label{fig:mainresults} 
\end{figure}
\vspace{-2mm}
\subsection{Face Synthesis with Flexible Modal Combinations}
\label{sec:application}
Our uni-modal training strategy with modal surrogates enables us to involve more modalities without considering their complex combination, and linear increase of inference complexity.
The entropy-aware modal-adaptive modulation further finely adjust noise according to given condition and modal-specific information, enabling various adaptive face synthesis.
Therefore we could efficiently integrate more modalities and achieve a wide range of face synthesis applications, shown in Fig. \ref{fig:teaser} and Fig. \ref{fig:mainresults}.

\begin{figure}[t] 
    \centering 
    \includegraphics[width=1\linewidth]{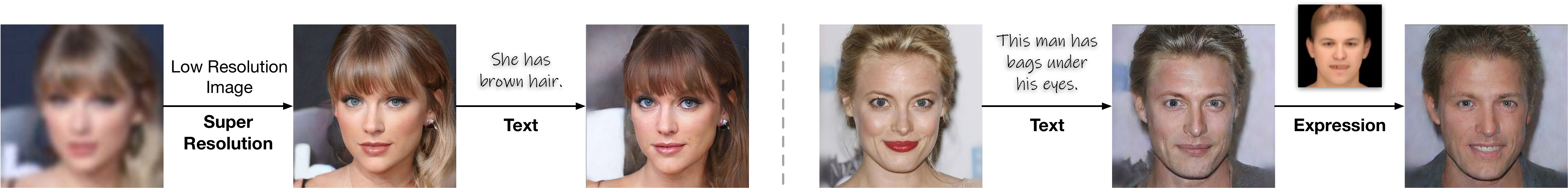} 
    \caption{ Our method's potential in integration with existing editing method\cite{kawar2023imagic} to achieve flexible editing tasks.} 
    \label{fig:editing} 
\end{figure}
\noindent
\textbf{Diverse uni-modal synthesis.}
Fig. ~\ref{fig:mainresults} (a) (left part) demonstrates our model's adeptness in uni-modal synthesis, consistently producing high-quality facial images with great fidelity to various single modal condition.
Whether the condition is face layout from mask, an expression from a 3D face model, or detailed structure in a sketch, our model captures these modal intrinsic accordingly and synthesis facial results of high fidelity, quality and diversity, within a single network.
The pleasing uni-modal results demonstrate that our modal surrogate learns its modal intrinsic via Eq. \ref{eq:uniloss1} and decorate condition with modal-specific information, therefore our model can discriminate various types of conditions and synthesis face accordingly.

\noindent
\textbf{Diverse combination of multi-modal synthesis.} 
Our method's integration of various modalities allows for versatile combinations in face synthesis. For example, we can create face according mask layout with lighting effect (Fig.~\ref{fig:mainresults} (b)) or based on poses and text (Fig.~\ref{fig:mainresults} (c)). Our approach achieve text-guided super-resolution (Fig. ~\ref{fig:mainresults} (d)), and combines sketches with lighting effects to generate face (Fig.~\ref{fig:mainresults} (e)). It can even merge lighting, text, sketch, and mask inputs to produce diverse, high-quality faces (Fig.~\ref{fig:mainresults} (f)), demonstrating our method's impressive flexibility and adaptivity in various facial image synthesis tasks.
Note that the synthesis results are obtained within a \textit{single sampling process of a single synthesis network} regardless of the number of given conditions.
We achieve face synthesis under flexible combination of multi-modal, no need to consider such complex modal combination during training. Instead we train our model in a uni-modal manner and our modal surrogate efficiently learns interaction among multi-modal conditions via Eq.~\ref{eq:uniloss2}.

\noindent
\textbf{Flexible multi-modal face editing.} 
Fig.~\ref{fig:editing} demonstrates our method's potential of integrating existing editing methods to achieve flexible  editing tasks.
We have incorporated editing method Imagic~\cite{kawar2023imagic} into our multi-modal face synthesis framework.
We begin by fixing the network parameters and optimizing the conditional embedding, which is initially set as the target condition feature. 
Next, the conditional is frozen and the diffusion U-Net parameters are further fine-tuned to reconstruct the image. The final step involves blending the feature embedding with the target embedding, guiding the diffusion network to generate the final results. 

\begin{table}[t]
\caption{Comparison with existing methods on multi-modal conditioned (mask + text) face synthesis.}
\label{tab:comparison}
\centering
\resizebox{0.7\linewidth}{!}{%
\begin{tabular}{cccccc}
\toprule 
Method &
 \cellcolor{color3}{ Training Data Type} &
  \# of Synthesis Network&
\cellcolor{color3}FID $\downarrow$ &
  Text(\%) $\uparrow$ &
\cellcolor{color3}Mask (\%) $\uparrow$
  \\ \midrule \midrule 
TediGAN \cite{xia2021tedigan}  &Uni-modal&1 &  116.04&  24.38&    86.37 \\ 
Compose \cite{liu2022compositional, nair2023unite} & Uni-modal& 2&127.39 &24.22  & 77.34    \\ 
CoDiff \cite{huang2023collaborative}  & Multi-modal&2 &122.51  &24.37  &  85.94   \\  
ControlNet \cite{zhang2023adding}  &Multi-modal & 1&136.41&25.70 &  85.44 \\ 
T2I-Adapter \cite{mou2023t2i}  &Multi-modal &1&139.82 &\textbf{26.11} & 79.84  \\ 
Ours  &Uni-modal & 1&\textbf{103.14} &24.70 & \textbf{90.16}   \\ 
\midrule 
\end{tabular}%
}
\vspace{-5mm}
\end{table}

\begin{figure}[t]
\setlength{\abovecaptionskip}{1pt}
\centering
        \begin{minipage}[b]{0.1\textwidth}
        \includegraphics[width=\textwidth]{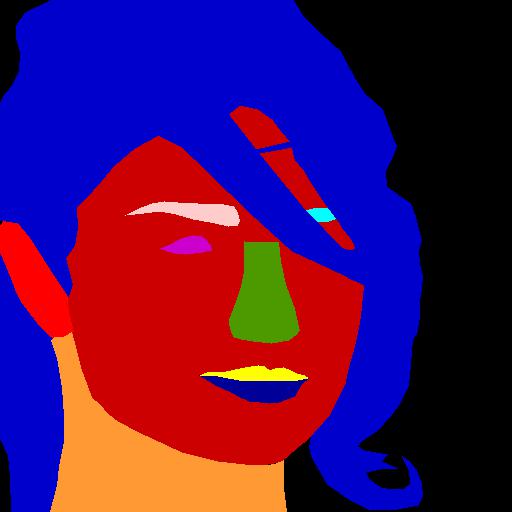}
    \end{minipage}
    \begin{minipage}[b]{0.15\textwidth}
        \tiny{
        \textit{The woman looks young. No beard on her face}}   
    \end{minipage}
     \begin{minipage}[b]{0.1\textwidth}
        \includegraphics[width=\textwidth]{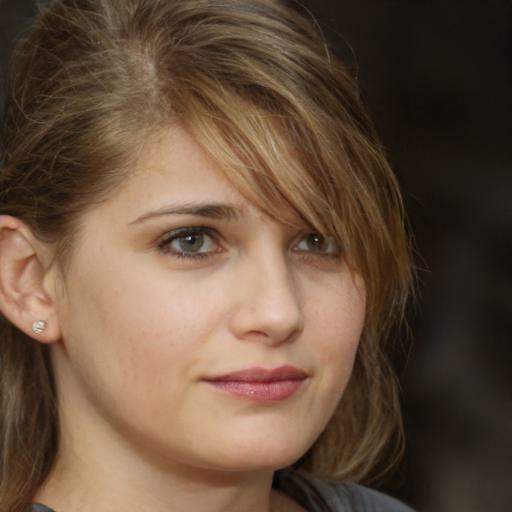}
        
    \end{minipage}
    \begin{minipage}[b]{0.1\textwidth}
        \includegraphics[width=\textwidth]{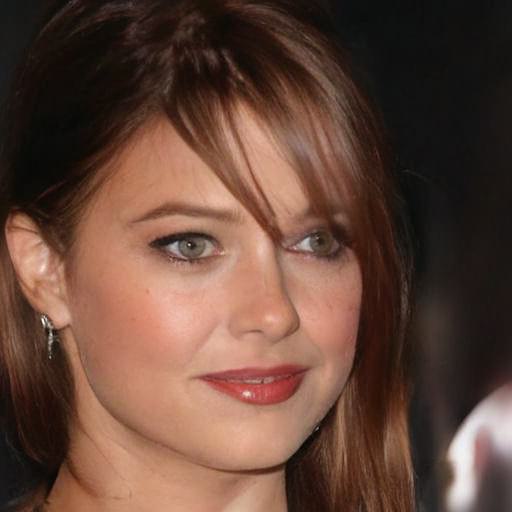}
         
    \end{minipage}
    \begin{minipage}[b]{0.1\textwidth}
        \includegraphics[width=\textwidth]{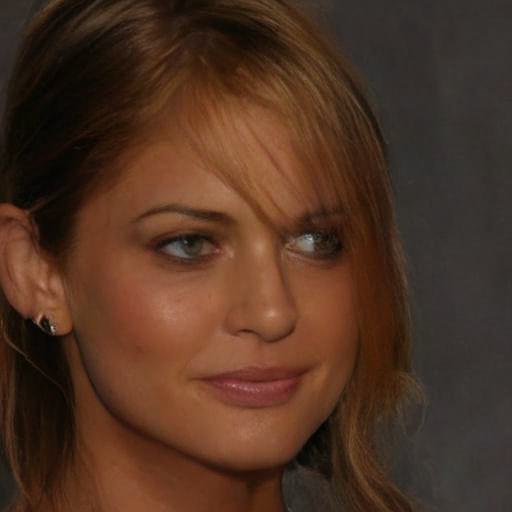}
        
    \end{minipage}
    \begin{minipage}[b]{0.1\textwidth}
        \includegraphics[width=\textwidth]{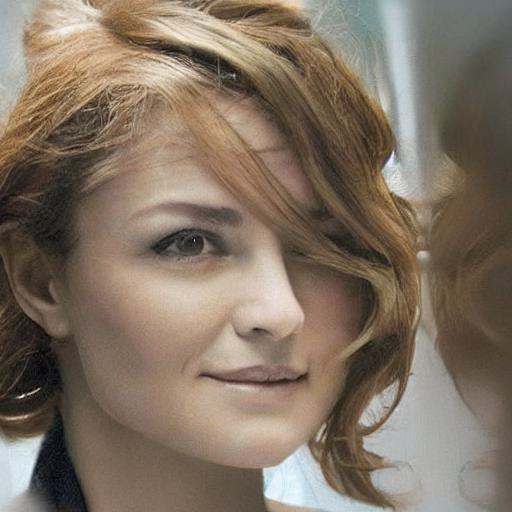}
       
    \end{minipage}
    \begin{minipage}[b]{0.1\textwidth}
        \includegraphics[width=\textwidth]{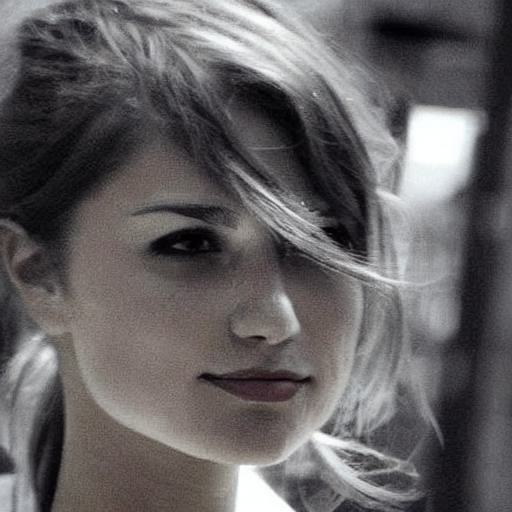}
        
    \end{minipage}
        \begin{minipage}[b]{0.1\textwidth}
        \includegraphics[width=\textwidth]{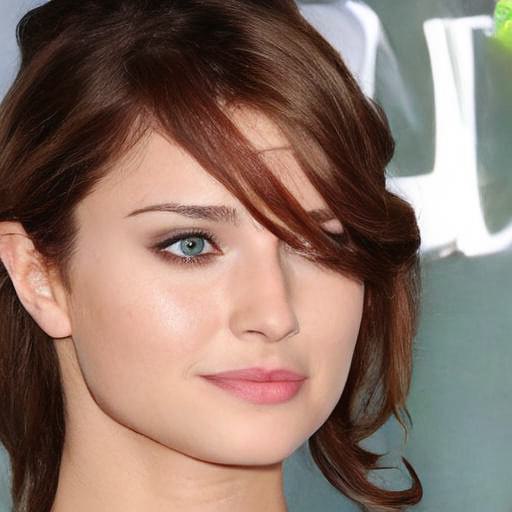}
    \end{minipage}
    \\
    \begin{minipage}[b]{0.1\textwidth}
        \includegraphics[width=\textwidth]{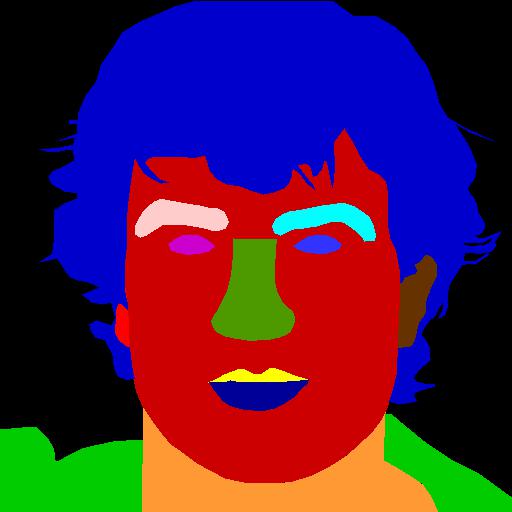}
       \caption*{\tiny Mask}
    \end{minipage}
    \begin{minipage}[b]{0.15\textwidth}
        \tiny{
        \textit{He is a teenager. His face is covered with his stubble.}}
        \caption*{\tiny Text}
    \end{minipage}
     \begin{minipage}[b]{0.1\textwidth}
        \includegraphics[width=\textwidth]{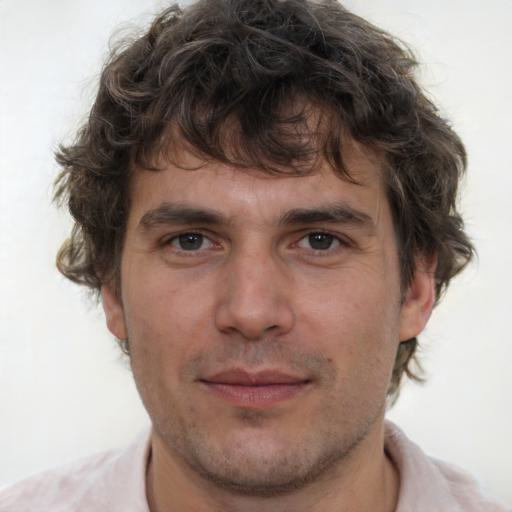}
        \caption*{\tiny TediGAN }
    \end{minipage}
    \begin{minipage}[b]{0.1\textwidth}
        \includegraphics[width=\textwidth]{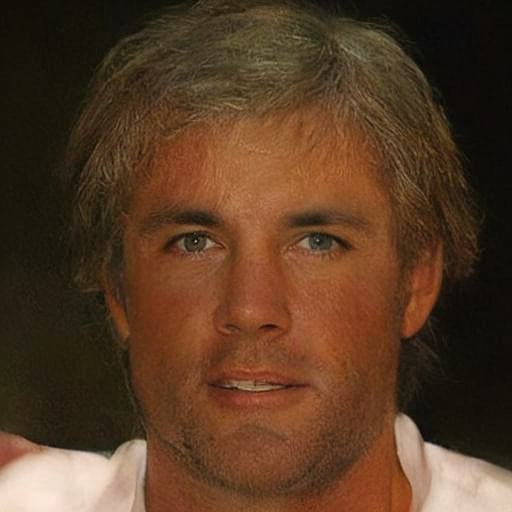}
        \caption*{\tiny Compose}
    \end{minipage}
    \begin{minipage}[b]{0.1\textwidth}
        \includegraphics[width=\textwidth]{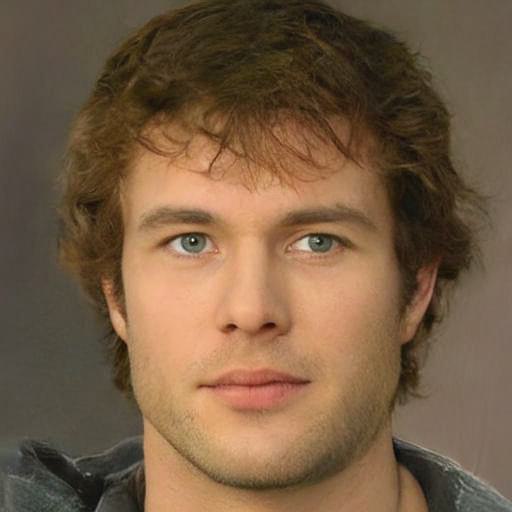}
        \caption*{\tiny CoDiff }
    \end{minipage}
    \begin{minipage}[b]{0.1\textwidth}
        \includegraphics[width=\textwidth]{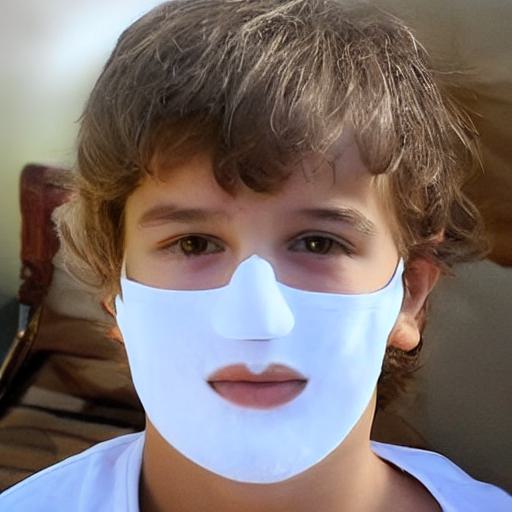}
        \caption*{\tiny {T2IAdapter }}
    \end{minipage}
    \begin{minipage}[b]{0.1\textwidth}
        \includegraphics[width=\textwidth]{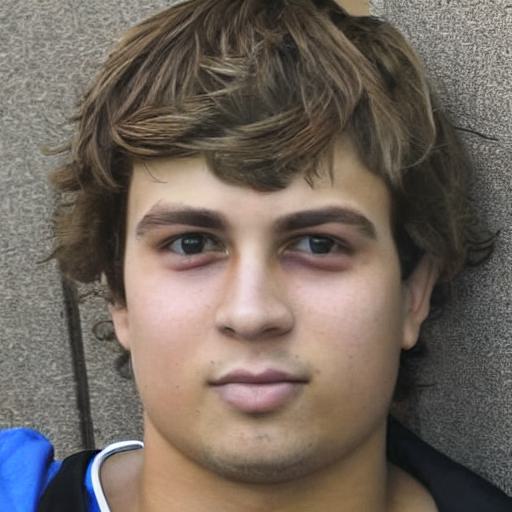}
        \captionsetup{font={small}}\caption*{\tiny ControlNet}
    \end{minipage}
        \begin{minipage}[b]{0.1\textwidth}
        \includegraphics[width=\textwidth]{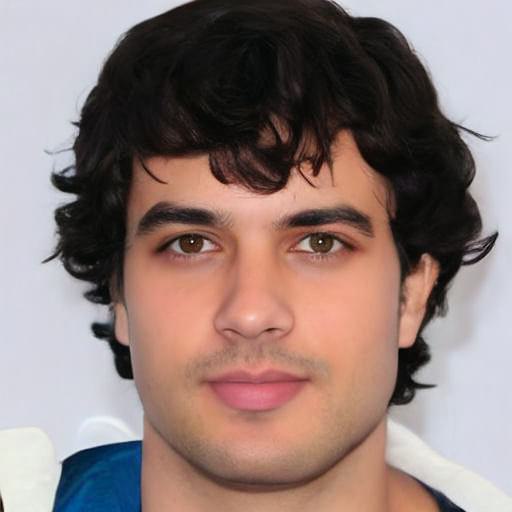}  
         \caption*{ \tiny { Ours} }
    \end{minipage}
\caption{Our methods generate face images of higher quality and align better with the given face parsing map and textual description, compared with existing competitive multi-modal condition synthesis methods.}
\label{fig:comparison_with_sota}
\end{figure}
  \vspace{-2mm}
\subsection{Comparison Analysis}
\label{sec:compare_study}
\textbf{Compared methods.}
We compare our method against several leading approaches: TediGAN \cite{xia2021tedigan}, Composable \cite{liu2022compositional}, Multimodal-diff \cite{nair2023unite}, Collaborative Diffusion \cite{huang2023collaborative}, ControlNet \cite{zhang2023adding}, and T2I-adapter \cite{mou2023t2i}. 

\begin{figure*}[t] 
    \centering 
    \includegraphics[width=0.8\linewidth]{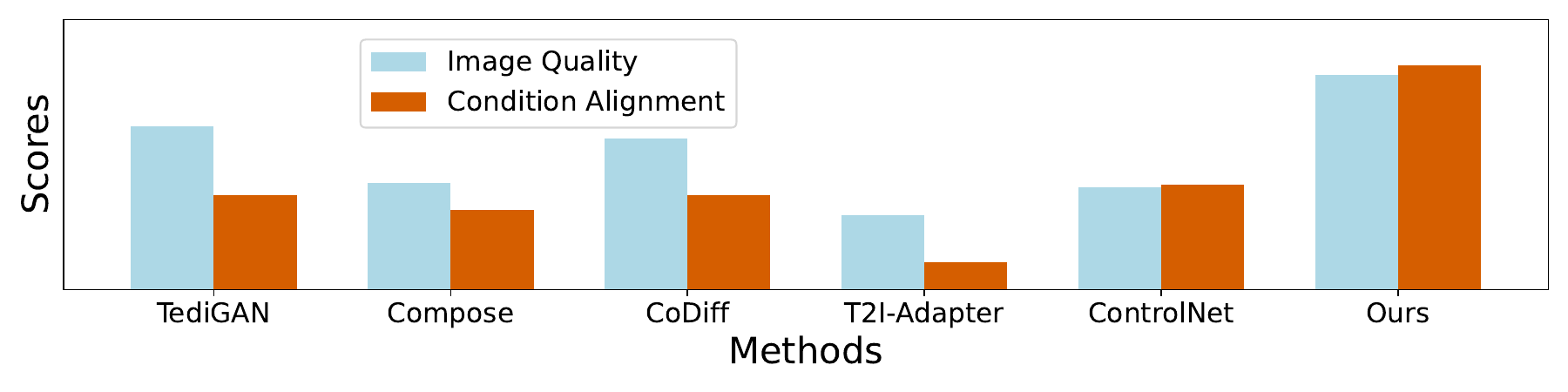} 
    \vspace{-2pt}
    \caption{User study. Our results achieve best among compared methods in terms with fidelity and quality of generated face.} 
    \label{fig:userstudy} 
\end{figure*}

\noindent\textbf{Quantitative and qualitative comparisons.} 
In Tab.~\ref{tab:comparison}, we benchmark our method's FID, Text, and Mask accuracy in multi-modal facial synthesis against leading techniques. 
TediGAN\cite{xia2021tedigan} yields high quality results by optimizing each condition and using a high-resolution generator StyleGAN. 
As shown in Fig. \ref{fig:comparison_with_sota}, however, TediGAN struggles in producing results of satisfying alignment and it involves per-instance fine-tuning for each condition.
The generated face either overlooks the mask detail (hair shape of woman) or neglects \textit{teenager} from text.
ControlNet \cite{zhang2023adding} and T2I-Adpter \cite{mou2023t2i} score well in text accuracy, leveraging advanced text encoders\cite{radford2021learning} and inheriting capabilities from Stable Diffusion\cite{rombach2022high}. 
As shown in Fig. \ref{fig:comparison_with_sota} and Tab. \ref{tab:comparison}, this line of methods requires multi-modal annotated data and produce noticeable artifacts.
Compose\cite{nair2023unite, liu2022compositional} lags behind due to its simplistic fusion of uni-modal model noise, overlooking crucial modality-specific and inter-modal collaboration in multi-modal synthesis. The generated face exhibits notable artifacts. 
Our method produce results of high quality that is align well with given conditions, e.g. the hair detail of given mask and the age and beard attribute from given text.
We achieve multi-modal face synthesis within a single synthesis network and require uni-modal annotate data under efficient uni-modal training.
\noindent\textbf{User study.}
We have invited 12 experienced researchers to score the results of compared methods, in terms of image quality and condition alignment. The higher the score, the better the results. Fig. \ref{fig:userstudy} shows that our method yields better performance in terms of image quality and condition alignment. 
\subsection{Ablation Analysis}
\label{sec:abla_anal}
We first validate how our uni-modal training with surrogate helps to achieve scalable and flexible multi-modal face synthesis. 
Next, we examine the effectiveness of our entropy-aware modal-adaptive modulation in fully leveraging the control of varing multi-modal conditions.

\noindent\textbf{Modal surrogate function: condition decoration.}
%
We demonstrate the effectiveness of modal surrogate to decorate condition. As shown in Tab. \ref{tab:abla_full}, the baseline $M_1$ first trains the parallel synthesis network for each modality and then fuses noise of each modal to achieve multi-modal face synthesis. $M_1$ produces acceptable results for the two uni-modal task (text/mask to face), but suffers from poor scalability for its linear increase for the parameter number and sampling complexity of the modal number.
$M_2$ improves $M_1$ by setting for each modality a trainable modal surrogate to decorate condition of its modal. The resulting network is capable of generating face under various type of conditions within a single model.
\begin{table}[t]
\caption{Ablation results of uni-modal training with modal surrogates and entropy-aware modal-adaptive modulation. For each task the top three are marked in \textcolor{red}{red}, \textcolor{blue}{blue}, and \textcolor{green}{green}, respectively.}
\label{tab:abla_full}
\centering
\resizebox{\linewidth}{!}{%
\begin{tabular}{c|ccc|ccc|ccc|ccc|ccc|ccc}
\hline
 & \multicolumn{3}{c}{$M_1$} & \multicolumn{3}{c}{$M_2$} &\multicolumn{3}{c}{$M_3$}& \multicolumn{3}{c}{$M_4$} &\multicolumn{3}{c}{$M_5$}&\multicolumn{3}{c}{$M_6$}\\
 \midrule \midrule 
\rowcolor{color3}\# of Model&
\multicolumn{3}{c}{Parallel model}&\multicolumn{3}{|c}{Shared model}&\multicolumn{3}{|c}{Shared model }&\multicolumn{3}{|c}{Shared model }&\multicolumn{3}{|c}{Shared model }&\multicolumn{3}{|c}{Shared model }\\
Training strategy&
\multicolumn{3}{c}{Uni-modal train}&\multicolumn{3}{|c}{Uni-modal train}&\multicolumn{3}{|c}{Uni-modal train}&\multicolumn{3}{|c}{Multi-modal train }&\multicolumn{3}{|c}{Multi-modal train}&\multicolumn{3}{|c}{Uni-modal train }\\
\rowcolor{color3}Surrogate&\multicolumn{3}{c}{-}&\multicolumn{3}{|c}{Condition decoration}&\multicolumn{3}{|c}{Condition decoration}&\multicolumn{3}{|c}{-}&\multicolumn{3}{|c}{Condition decoration}&\multicolumn{3}{|c}{Condition decoration }\\
\rowcolor{color3}function&\multicolumn{3}{c}{-}&\multicolumn{3}{|c}{-}&\multicolumn{3}{|c}{Inter-modal learning}&\multicolumn{3}{|c}{-}&\multicolumn{3}{|c}{-}&\multicolumn{3}{|c}{Inter-modal learning}\\
Adjust Noise&\multicolumn{3}{c}{-}&\multicolumn{3}{|c}{-}&\multicolumn{3}{|c}{-}&\multicolumn{3}{|c}{-}&\multicolumn{3}{|c}{-}&\multicolumn{3}{|c}{\checkmark}\\ 
\midrule
\rowcolor{color3}& FID $\downarrow$ &Text $\uparrow$  & Mask$\uparrow$ &
FID $\downarrow$ &Text$ \uparrow$  & Mask$\uparrow$ &
FID $\downarrow$ &Text$ \uparrow$  & Mask$\uparrow$ &
FID $\downarrow$ &Text$ \uparrow$  & Mask$\uparrow$ &
FID $\downarrow$ &Text$ \uparrow$  & Mask$\uparrow$ &
FID $\downarrow$ &Text$ \uparrow$  & Mask$\uparrow$ \\
Mask to Face & 113.77&-&87.21 &\textcolor{blue}{109.31}&-&\textcolor{teal}{89.27} &\textcolor{teal}{109.83}&-&88.97&-&-&-&\textcolor{red}{108.69}&-&\textcolor{blue}{90.06} &112.91&-&\textcolor{red}{90.31} \\
\rowcolor{color3}Text to Face &\textcolor{red}{108.81}&\textcolor{blue}{24.54}&-&\textcolor{teal}{109.61}&24.29&-&114.07&\textcolor{teal}{24.48} &-&-&-&-&195.94&23.18 &-&\textcolor{blue}{110.32}&\textcolor{red}{24.78} &-\\
Mask\&Text to Face  & 117.94&24.24&78.55 &121.77&24.07&88.33 &\textcolor{teal}{113.29}&\textcolor{teal}{24.28}&\textcolor{teal}{89.72} &124.98&\textcolor{teal}{24.28}&85.44&\textcolor{blue}{112.25}& 
\textcolor{blue}{24.42}&\textcolor{blue}{89.91} &\textcolor{red}{103.14}&\textcolor{red}{24.70}&\textcolor{red}{90.16}\\
\hline
\end{tabular}
}

\end{table}
\begin{figure*}[!h] 
    \centering 
    \includegraphics[width=1\linewidth]{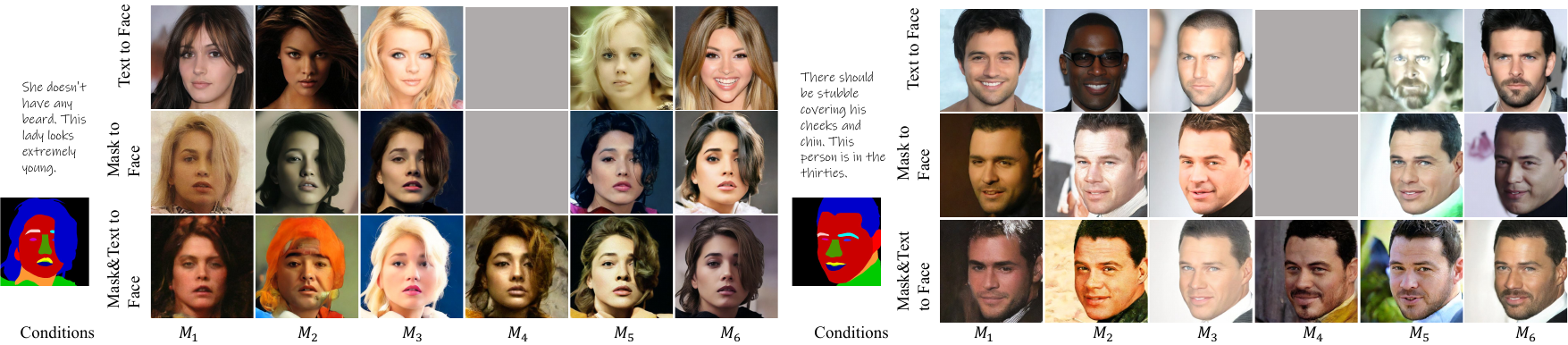} 
    \caption{Comparison of synthesis results across training methodologies. Our uni-modal training with modal surrogate enables a flexible and scalable face synthesis framework. Entropy-aware Modality-Adaptive Modulation further enhances the fidelity to given conditions. } 
    \label{fig:abla_full} 
\end{figure*}
\begin{figure*}[!htbp] 
    \centering 
    \includegraphics[width=1\linewidth]{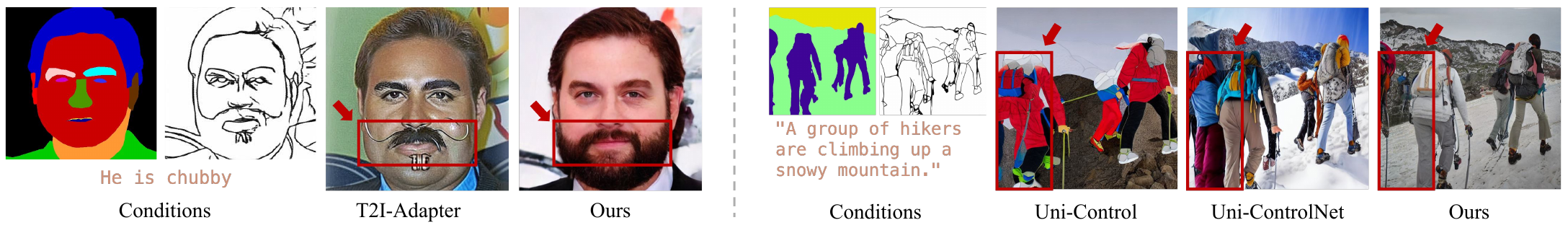} 
    \caption{Further verification of our method's effectiveness to learn inter-modal collaboration for multi-modal synthesis. T2I-Adapter and Uni-control directly combine features from multi-modal layout, and they do not modelling inter-modal collaboration.} 
    \label{fig:furter_surrogate} 
\end{figure*}

\noindent\textbf{Modal surrogate function: inter-modal learning.}
We demonstrate the effectiveness of modal surrogate to learn inter-modal interaction in helping achieve multi-modal face synthesis.
Both $M_1$ and $M_2$ produce poor results for multi-modal synthesis, shown in Tab. \ref{tab:abla_full} and Fig. \ref{fig:abla_full}. That is because their training process do not involve how multi-modal conditions interact in face synthesis process.
$M_3$ improves $M_2$ by extra involving all other modal surrogates during training. 
In this way the surrogates of some modal learned from data of other modalities, and thus serve as inter-modal linker for multi-modal synthesis.
Tab. \ref{tab:abla_full} and Fig. \ref{fig:abla_full} demonstrate that compared with $M_2$, $M_3$ supports multi-modal synthesis (text \& mask to face) of higher quality and condition alignment.
We further compare the method of combining uni-modal noise or features and our method in multi-modal synthesis in Fig. \ref{fig:furter_surrogate}. Directly combining uni-modal feature or noise tends to yield poor results as it lacks of multi-modal modelling in multi-modal synthesis. In contrast, our method effectively fuses constraints from multiple layouts.
Our cross-modal updating of surrogates helps to grasp multi-modal interaction with efficient uni-modal training.

\noindent\textbf{Uni-modal Training}
Our uni-modal training helps network avoid the short cut issue in multi-modal learning, and fully learns synthesis for each modality.
$M_5$ represents the method of multi-modal training with modal surrogate as condition decoration.
Tab. \ref{tab:abla_full} and Fig. \ref{fig:abla_full} demonstrate that $M_5$ tends to rely more on mask and overlook text condition, as the performance of text synthesis is rather bad.
In contrast our uni-modal training strategy $M_3$ fully learns synthesis under each modality condition, thus resulting flexible synthesis versatile across all modalities.
\begin{figure*}[t] 
    \centering 
    \includegraphics[width=\linewidth]{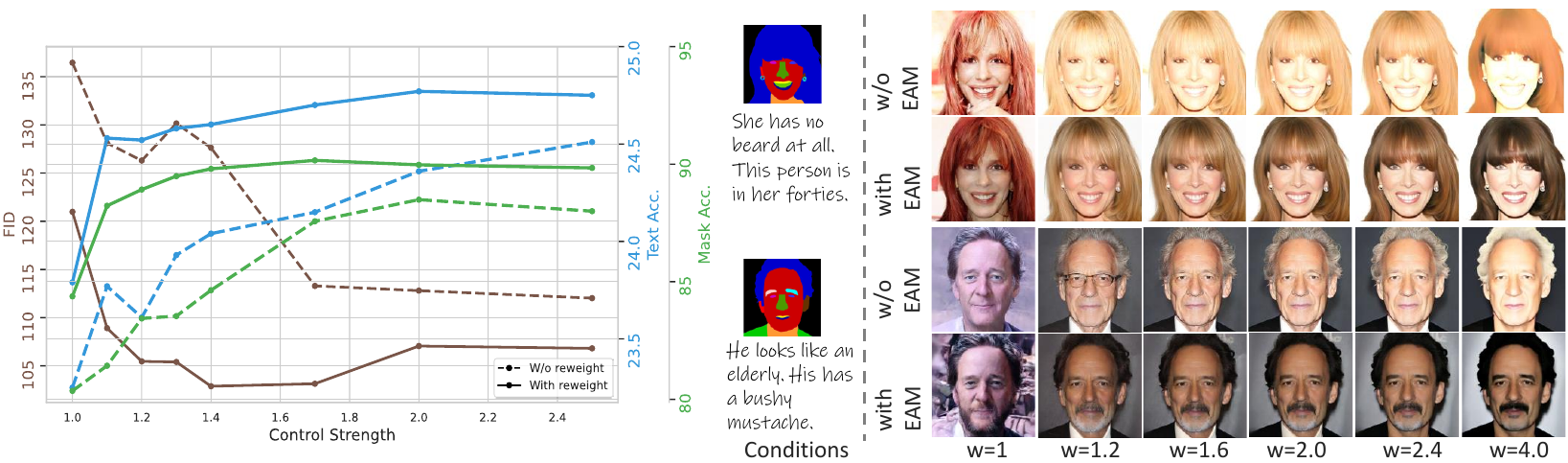} 
     \vspace{-2mm}
    \caption{Quantitative and qualitative comparison of synthesis results of our full methods and ours without Entropy-aware Modal-adaptive Modulation (EAM) under different control strength. As control strength increases, the fidelity of synthesis results increases and our full methods achieve consistently superior results than ours without EAM.} 
    \label{fig:abla_reweighting} 
\end{figure*}
   
\noindent\textbf{Entropy-aware modal-adaptive modulation.}
Building upon uni-modal training with modal surrogate $M_3$, we further integrate entropy-aware modal-adaptive modulation $M_6$ to adjust noise level in accordance with the given condition.
Fig.~\ref{fig:abla_full} shows that $M_3$ neglect some constraint of multi-modal condition, \eg, the man face has no beard. $M_3$ tends to overlook conditions of high entropy (text) for multi-modal synthesis.
$M_6$ shows improvement in terms with condition alignment of both conditions for both uni-modal and multi-modal synthesis.
In the right part of Fig.~\ref{fig:abla_reweighting} (w/o EAM) shows that even finely adjusting the user-defined guidance weight $w$ of CFG \cite{ho2022classifier} still fails to obtain results of satisfying quality and alignment. 
In contrast, our full method (with EAM) exhibits high fidelity and image quality due to our finely, real-time adjustment of the de-noising level based on the conditions and the current state of de-noising. 

\vspace{-1mm}
\section{Conclusion}
\vspace{-3mm}
In conclusion, our research presents a significant advancement in the field of multi-modal face synthesis, presenting a highly scalable framework that supports face synthesis of high quality and fidelity under flexible combination of condition.
Our approach introduces a uni-modal training framework with modal surrogates for each modality that serve as condition decorator for its modality and an inter-linker to facilitate inter-modal collaboration.
%
The entropy-aware modal-adaptive modulation precisely tunes diffusion noise based on modal characteristics and conditions, enhancing the de-noising process for superior synthesis quality.
%
Our method broadens multi-modal face synthesis capabilities, supporting a wide range of synthesis tasks from uni-modal to complex multi-modal combinations.

%
%
\bibliographystyle{splncs04}
\bibliography{main}
\end{document}